\documentclass[letterpaper, 10 pt, journal, twoside]{IEEEtran}  
\IEEEoverridecommandlockouts          
\usepackage[pdftex]{graphicx}
\graphicspath{{../pdf/}{../jpeg/}}
\DeclareGraphicsExtensions{.pdf,.jpeg,.png}

\usepackage{algorithm}
\usepackage{amsmath}
\usepackage[noend]{algpseudocode}
\usepackage{euscript}
\usepackage{color}
\usepackage{booktabs} 

\makeatletter
\let\NAT@parse\undefined
\makeatother
\usepackage[numbers,sort,compress]{natbib}
\usepackage{amsfonts}
\usepackage{mathtools}
\usepackage{placeins}
\usepackage{microtype}
\usepackage{soul}
\usepackage{tikz}
\usepackage{todonotes}
\usepackage{booktabs}	

\usepackage[caption=false,font=footnotesize]{subfig}
\usepackage[hidelinks]{hyperref}
\usepackage[capitalise]{cleveref}
\usepackage{fancyhdr}

\crefformat{section}{#2Sec.~{\textup{#1}}#3}
\crefformat{figure}{#2Fig.~{\textup{#1}}#3}
\crefformat{equation}{#2({\textup{#1}})#3}
\crefformat{table}{#2Table~{\textup{#1}}#3}
\crefmultiformat{figure}{Figs.~#2#1#3}{ and~#2#1#3}{}{}
\crefmultiformat{section}{Secs.~#2#1#3}{ and~#2#1#3}{}{}
\crefrangeformat{section}{Secs.~#3#1#4 to~#5#2#6}

\newcommand{\transform}[1][]{%
	\ifthenelse{\equal{#1}{}}{\mathbf{T}}{\mathbf{T}_{#1}}%
}

\usepackage[scr=boondox]{mathalfa}

\newcommand{\se}[1][3]{$SE\left(#1\right)$}
\newcommand{\dataurl}{\url{https://robotic-esp.com/datasets/}}

\title{\LARGE \bf
The Oxford Multimotion Dataset:\\
Multiple SE(3) Motions with Ground Truth
}

\author{Kevin M. Judd$^{1}$ and Jonathan D. Gammell$^{1}$%
\thanks{Manuscript received: September, 10, 2018; Revised December, 3, 2018; Accepted January, 4, 2018.}
\thanks{This paper was recommended for publication by Editor Eric Marchand upon evaluation of the Associate Editor and Reviewers' comments.}
\thanks{$^{1}$K. M. Judd and J. D. Gammell are with the Estimation, Search, and Planning (ESP) research group at the Oxford Robotics Institute (ORI), University of Oxford, United Kingdom. {\tt\footnotesize \{kjudd, gammell\}@robots.ox.ac.uk}}%
\thanks{$^{2}$The dataset and associated tools are available at \dataurl}%
\thanks{Digital Object Identifier (DOI): see top of this page.}
}

\begin{document}

\maketitle

\begin{abstract}
Datasets advance research by posing challenging new problems and providing standardized methods of algorithm comparison.
High-quality datasets exist for many important problems in robotics and computer vision, including egomotion estimation and motion/scene segmentation, but not for techniques that estimate \emph{every} motion in a scene. 
Metric evaluation of these \emph{multimotion} estimation techniques requires datasets consisting of multiple, complex motions that also contain ground truth for \emph{every} moving body.

The \emph{Oxford Multimotion Dataset} provides a number of multimotion estimation problems of varying complexity. 
It includes both complex problems that challenge existing algorithms as well as a number of simpler problems to support development. 
These include observations from both static and dynamic sensors, a varying number of moving bodies, and a variety of different 3D motions.
It also provides a number of experiments designed to isolate specific challenges of the multimotion problem, including rotation about the optical axis and occlusion.\looseness=-1

In total, the Oxford Multimotion Dataset contains over 110 minutes of multimotion data consisting of stereo and RGB-D camera images, IMU data, and Vicon ground-truth trajectories. 
The dataset culminates in a complex toy car segment representative of many challenging real-world scenarios.
This paper describes each experiment with a focus on its relevance to the multimotion estimation problem. 
\end{abstract}


\section{Introduction}\label{sec:introduction}
\IEEEPARstart{T}{he} field of robotic vision has seen significant progress in the areas of segmentation, reconstruction, estimation, and tracking.
Much of this progress has been stimulated by the publication of diverse, specialized datasets. 
These datasets can both essentialize a problem, isolating a specific task from the many challenging aspects of visual analysis, as well as create a meaningful reference with which to evaluate different techniques. 
Well-cultivated datasets reduce the burden of evaluating novel methods and accelerate the advancement of state-of-the-art techniques.

\begin{figure}[t!]
	\centering%
	\vspace{1mm}
	\subfloat[\label{fig:marquee:image}]{%
		\includegraphics[width=\columnwidth]{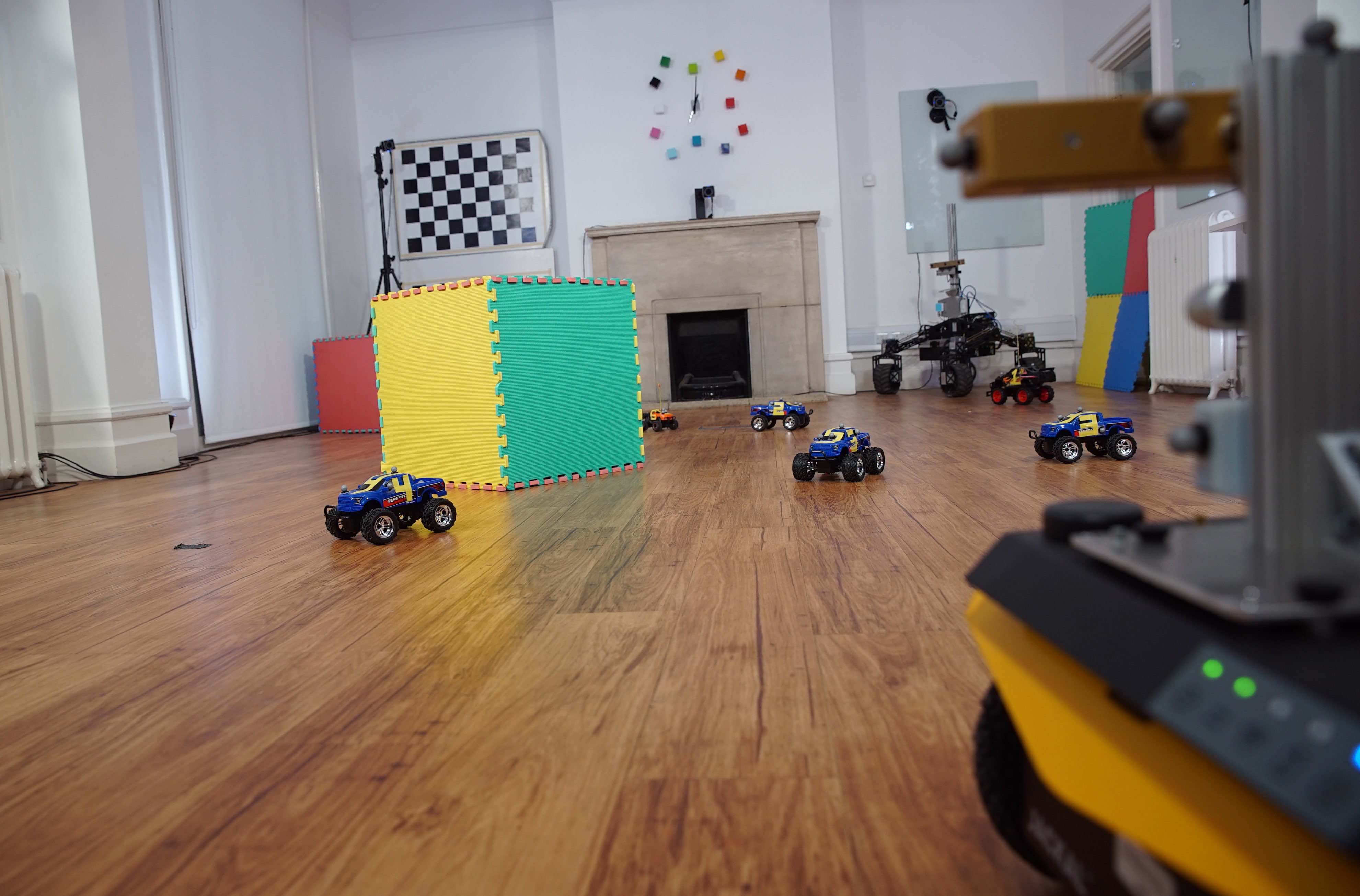}%
	}
	
	\subfloat[\label{fig:marquee:plot}]{%
		\includegraphics[clip,width=\columnwidth]{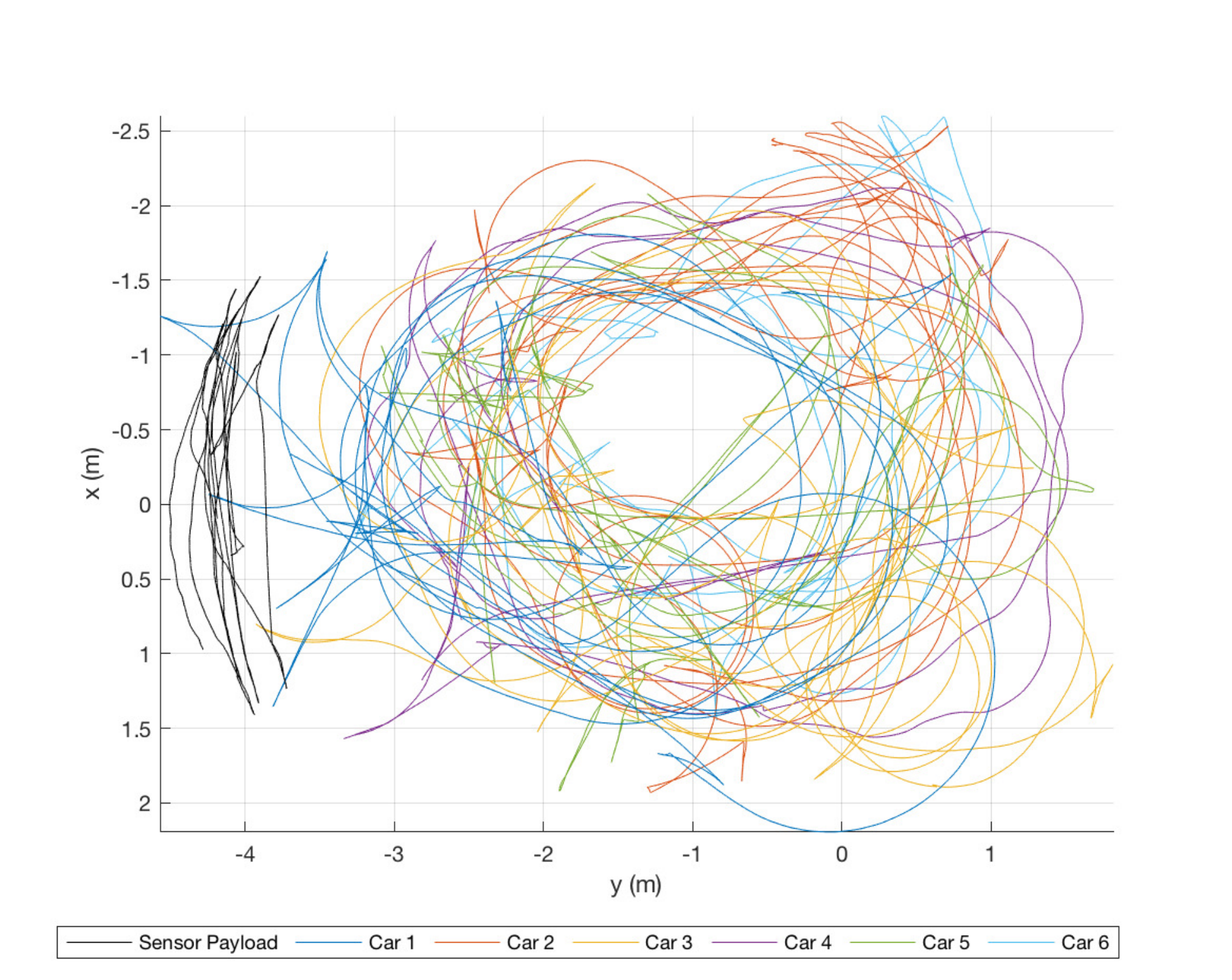}%
	}%
	\caption{One segment of the dataset consisting of six radio-controlled toy cars (a, background) observed by a robot-mounted sensor platform (a, foreground). Color image, depth, and IMU streams from the sensor platform are included in the dataset, as well as the Vicon ground-truth trajectories of all motions in the scene (b). The camera was mounted on a Clearpath Robotics Jackal platform and its motion is shown in black. The motions of the toy cars in the scene are shown in other colors.}\label{fig:marquee}%
\end{figure}

A variety of datasets \cite{sturm2012, geiger2012, tron2007, ochs2014, gaidon2016, dosovitskiy2017} now exist to address the individual challenges of scene segmentation, pose estimation, and object tracking, but none are well-suited to address the \emph{multimotion estimation} problem.
As an extension of egomotion estimation which focuses on estimating the motion of a camera through its environment, multimotion estimation requires both segmentation and estimation in order to estimate 6-DOF, i.e., \se[3]{}, trajectories for every motion in a scene.
To lay the foundation for addressing the multimotion estimation problem, a new dataset is required that includes ground-truth trajectory information for every moving object within a scene.

This paper introduces the Oxford Multimotion Dataset$^{2}$, which consists of over 110 minutes of stereo and RGB-D camera images and IMU data, as well as ground-truth trajectory information for both the camera egomotion and every independently moving object within each scene. 
The dataset consists of three primary scenes (\cref{sec:dataset:primary}), each with static, mostly translational (i.e., mostly $\mathbb{R}^{3}$), and unconstrained $SE\left(3\right)$ sensor motions.
These thirteen different data segments highlight specific milestones and targets in the progression toward fully addressing the multimotion estimation problem. 
They are accompanied by additional experiments and data segments that focus on particularly interesting motions (\cref{sec:secondary}). 

The first scene of the dataset, \textit{Swinging Blocks} (\cref{sec:dataset:swinging}), features four independently swinging blocks observed with static, $\mathbb{R}^{3}$, and $SE\left(3\right)$ sensor motions and is designed to test the ability of an algorithm to estimate multiple, full \se[3]{} motions.
The second scene, \textit{Occluding Blocks} (\cref{sec:dataset:occlusion}), involves a swinging block which is partially and fully occluded by a tower of blocks sliding in front of it.
The objects are also observed with static, $\mathbb{R}^{3}$, and $SE\left(3\right)$ sensor motions and it is designed to test the ability of an algorithm to handle missing information and extrapolate estimates through occlusions.
These two scenes isolate specific aspects of the multimotion estimation problem and lead toward the final scene, \textit{Toy Cars} (\cref{sec:dataset:cars}).
This scene involves several different experiments with up to six individual radio-controlled toy cars driven around a room with a static occlusion.
Data are collected with static, $\mathbb{R}^{3}$, and $SE\left(3\right)$ sensor motions, as well as from a wheeled robotic platform.
This robot perspective poses a difficult multimotion estimation problem representative of the complex challenges we believe robotics need to be able to solve.

\section{Background} \label{sec:review}
As described in \cite{judd2018}, the multimotion estimation problem seeks to estimate every motion in a scene, including that of the camera. 
This involves \emph{simultaneously} estimating and segmenting the motions, as each observation is assigned to a motion and each motion is estimated from those measurements.
Motions may be represented as sequences of discrete \se[3]{} transforms or continuous-time trajectories.

This dataset supports the development and testing of multimotion estimation algorithms by providing experiments with multiple rigid-body motions and ground truth. 
Techniques developed on these experiments will likely also be applicable to scenes consisting of piecewise-rigid motions.

\subsection{Existing Datasets}
Several datasets exist to address \emph{aspects} of the multimotion estimation problem, specifically for motion segmentation and egomotion estimation. They are each well-suited for their intended task but are unsuitable for extension to the multimotion estimation problem.

The TUM RGB-D SLAM dataset \cite{sturm2012} is a large collection of RGB-D image sequences and ground-truth egomotion information.
The RGB-D images were collected using a Microsoft Kinect, and the motion of the sensor was obtained from a Vicon motion capture system.
The dataset includes a variety of \se[3]{} camera motions and even several dynamic scenes, but there is no ground-truth information available for the other motions in the scene.

The KITTI autonomous driving dataset \cite{geiger2012} is a suite of different benchmarks addressing various computer vision challenges involved in autonomous driving, such as visual odometry (VO) and pedestrian tracking. 
The data was collected using a car-mounted sensor platform that includes a stereo camera, a 360$^{\circ}$ laser scanner, and a GPS. 
Although the dataset includes several complex urban scenes with multiple dynamic motions, there is no ground-truth information available for those motions. 
Furthermore, the dataset is tailored toward autonomous driving scenarios and most motions are predominantly \se[2]{}.

The Hopkins 155 dataset \cite{tron2007} consists of 155 monocular image sequences with two or three independent motions in each. 
The majority of the sequences involve checkerboard-patterned objects in an indoor environment, but others involve traffic scenes or other piecewise-rigid and non-rigid motions. 
The key contribution is the inclusion of tracked feature points through each sequence as well as the ground-truth segmentation of those features. 
The dataset has been used extensively as a motion segmentation benchmark, but it is not appropriate as a multimotion estimation dataset due to the lack of ground-truth trajectory information. 
Additionally, full \se[3]{} motions cannot be estimated from monocular images alone.

Part of the Hopkins 155 dataset was incorporated into the Freiburg-Berkeley Motion Segmentation Dataset \cite{ochs2014}, which contributes dense, pixel-wise annotations of each moving object in 59 different image sequences. 
The dataset remains unsuitable for multimotion estimation, as it still does not include any ground-truth trajectory information.

Synthetic datasets (e.g., \cite{gaidon2016} and \cite{dosovitskiy2017}) simulate autonomous driving environments with high levels of photorealism and customization. 
These frameworks are useful for developing data-driven computer-vision techniques, but do not replace the need for real-world data, especially for estimation approaches.

\subsection{Related Techniques}
A variety of research has addressed individual subsets of the motion estimation problem, particularly in multiple object tracking and multibody structure from motion. 
Unfortunately, the majority of these techniques fall short of completely addressing the multimotion estimation problem in \se[3]{}. 

Optical flow \cite{horn1981} and scene flow \cite{vedula1999} techniques find the 2D or 3D velocity vector of each point in a scene. 
These velocities can then be clustered into similar motions \cite{lenz2011} or segmented along discontinuities \cite{menze2015}.
The velocities are inherently translational and motions with significant rotation (i.e., \se[3]{}) require three or more points to estimate.
Estimating these motions requires segmenting the pixel-wise flow vectors into groups with similar \se[3]{} motions.

Appearance-based tracking detects \emph{objects} in images, associates the detections through time, and estimates the object motions using these chains of detections \cite{milan2016}. 
These object detectors are often highly specialized for specific objects or classes of objects, making them limited in application or fragile to appearance changes and requiring refinement over time \cite{kalal2012}. 
Additionally, the object detections generally exist as positions in $\mathbb{R}^3$, meaning they do not fully encapsulate the \se{} motions of the objects.
To fully address the multimotion estimation problem, these tracking techniques must be able to track any arbitrary object and estimate its \se[3]{} motion. 

Factorization methods use the affine camera model to project the motions in the scene into lower-dimensional subspaces. 
The affine model approximates the nonlinear perspective projection with a linear parallel projection. 
This simplifies the camera model but introduces severe projection errors in scenes with a wide field of view or a large depth of field \cite{hartley2003}. 
Coistera and Kanade \cite{costeira1998} use the affine model and matrix factorization to decompose tracked image points into the motion and shape of each object in a scene. 
In addition to the affine limitations, these factorization techniques generally require features to be successfully tracked throughout the entire estimation window. 
This is a difficult constraint to fulfill, especially in highly dynamic scenes with significant occlusion.
Some progress has been made in relaxing these constraints \cite{vidal2004, li2007, vidal2003}, but further work is required to fully address the multimotion estimation problem.\looseness=-1

Sampling techniques, such as RANSAC \cite{fischler1981}, estimate and fit a motion model to data in the presence of noise. 
Multiple models can be found by generating and merging a large number of initial hypotheses \cite{schindler2006} or by fitting and removing the dominant model in the data and recursively searching the outliers for the next model \cite{torr1998}.
As the signal-to-noise ratio decreases, the likelihood of sampling data points from a single, consistent model decreases.
This makes finding a motion in complex dynamic scenes challenging because all other motions are outliers that decrease the signal-to-noise ratio and make it harder to find correct models.

Energy minimization approaches segment data into multiple labels simultaneously by reducing a cost function. 
In multimotion estimation, this cost is designed to encompass how well the estimated trajectories describe the data, e.g., reprojection or photoconsistency error, as well as encourage piecewise smoothness throughout the scene. 
Isack and Boykov \cite{isack2012} apply an $\alpha$-expansion and model-refitting framework to segment motions by first sampling the data to estimate a large number of motion models (similarly to \cite{schindler2006}) and then refining the models and segmentation. 
Judd et al. \cite{judd2018} introduce Multimotion Visual Odometry (MVO), which uses energy minimization to extend the traditional VO pipeline to simultaneously segment and estimate the \se{} trajectory of every motion in the scene.

\begin{figure*}[t]
	\centering%
	\subfloat[\label{fig:apparatus:image}]{%
		\includegraphics[width=.95\columnwidth]{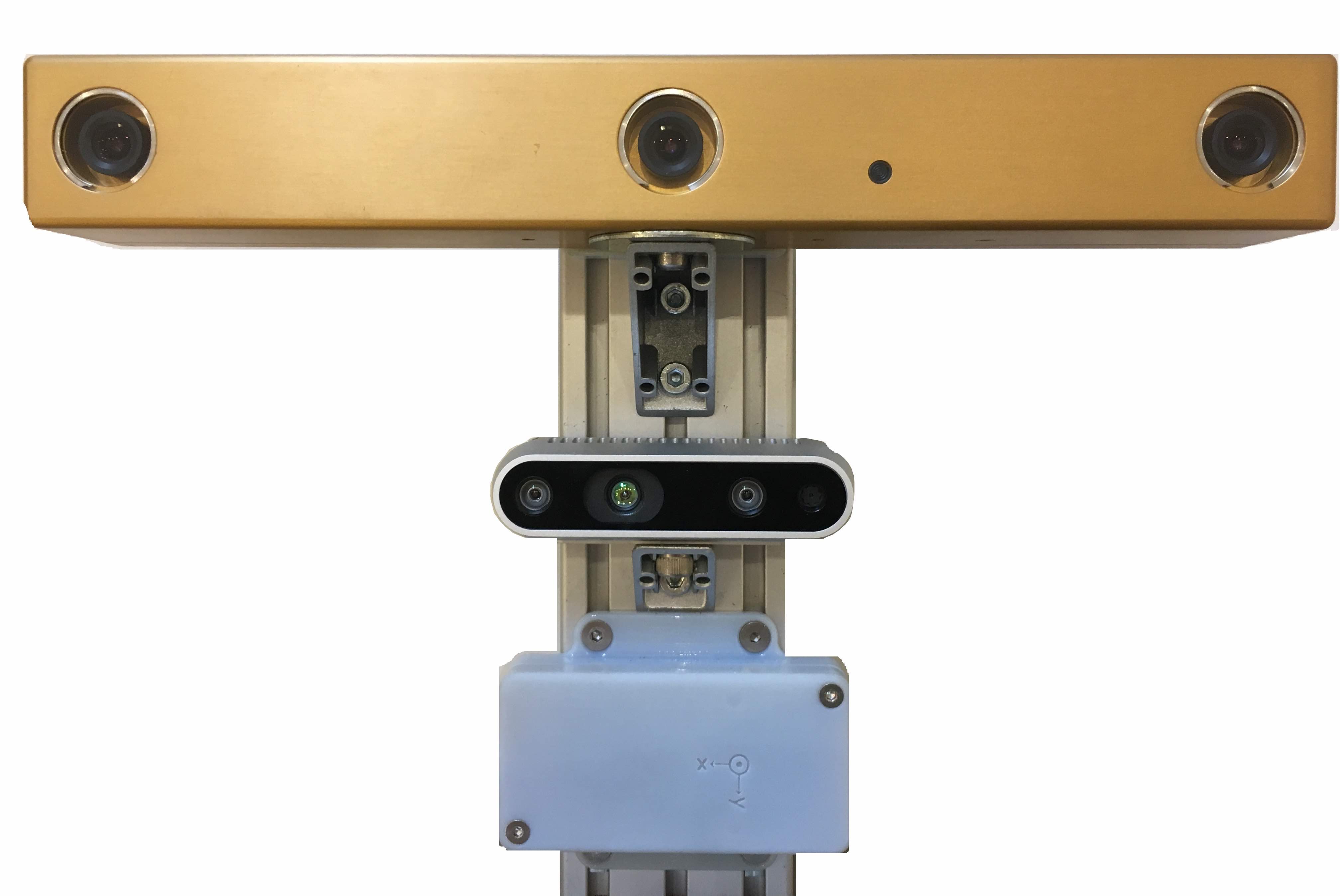}%
	}\hfill
	\subfloat[\label{fig:apparatus:diagram}]{%
		\includegraphics[width=.85\columnwidth]{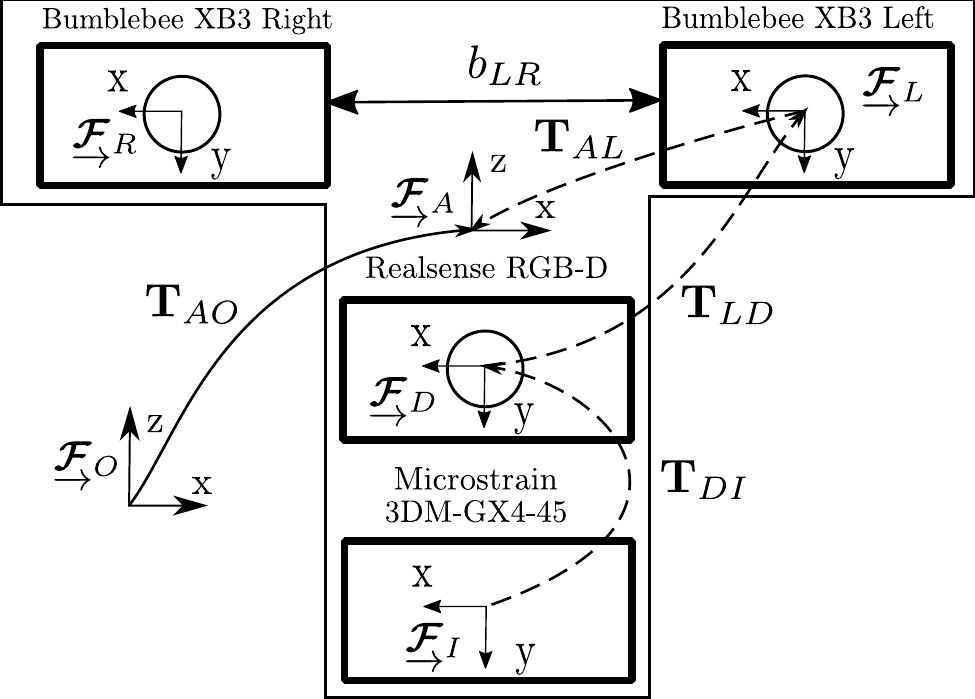}%
	}%
	\caption{The sensor apparatus (a) and calibration diagram (b) used in collecting the Oxford Multimotion Dataset. A Bumblebee XB3 provided calibrated stereo images, an Intel Realsense D435 RGB-D camera provided both color and depth images, and a Lord Microstrain 3DM-4GX-45 IMU provided orientation and acceleration information. The baseline between the left and right Bumblebee XB3 cameras, $b_{RL}$, is given by the manufacturer, and the transform between the Vicon origin frame and the Vicon frame for the apparatus, $\transform[AO]{}$, is provided by the Vicon motion capture system as ground truth (solid lines). The calibration between the depth camera and the left stereo camera, $\transform[LD]{}$, between the IMU and the depth camera, $\transform[DI]{}$, and between the left stereo camera and the Vicon apparatus frame, $\transform[AL]{}$, are calculated from the calibration sequences included in the dataset (dashed lines). All sensors are oriented with their z-axis forward and x-axis right, and the Vicon frames are aligned with the z-axis up and x- and y-axes arbitrary.}  	\label{fig:apparatus}
	\vspace{-1mm}
\end{figure*}

\vspace{2ex}
The majority of the existing motion estimation techniques fail to fully address the multimotion estimation problem. 
The Oxford Multimotion Dataset seeks to foster progress toward addressing this challenge by providing experiments designed for the development and testing of full-scene motion estimation. 
These include stereo and RGB-D camera images, IMU data, and Vicon ground truth for \emph{every} motion in the scene.

\section{Dataset} \label{sec:dataset}
The dataset was collected from a single, statically calibrated sensor platform in an indoor experimental Vicon room equipped with professional flicker-free lighting. 
The sensor apparatus consisted of a Point Grey Bumblebee XB3 stereo camera, an Intel Realsense D435 RGB-D camera, and a Microstrain 3DM-GX4-45 IMU (\cref{fig:apparatus}).
All sensor data was logged and timestamped by a single computer using ROS \cite{quigley2009}. 
The Bumblebee XB3 records synchronized stereo image pairs at 16 Hz and the Realsense records RGB images and 16-bit aligned depth images at 30 Hz.
The Microstrain IMU records orientation and acceleration data at 500 Hz (\cref{tab:sensors}).
This sensor combination lends itself readily to a variety of visual motion estimation methods, including stereo and RGB-D VO and visual-inertial odometry (VIO).

Ground-truth trajectory information for every motion in each scene, including the camera egomotion, was collected using a Vicon motion capture system. 
A Vicon coordinate frame is assigned to each object and its position and orientation are recorded at 200Hz.
The extrinsic calibrations between each sensor, as well as the Vicon system, are given in the dataset (\cref{sec:dataset:calibration}), along with the segments from which they are calculated (\cref{sec:secondary:calibration}). 

The dataset contains a series of increasingly difficult experiments designed to foster development of multimotion estimation algorithms. 
There are three primary sections (\cref{sec:dataset:primary}) of the dataset, each of which includes static and dynamic sensor segments in $\mathbb{R}^{3}$ and $SE\left(3\right)$.
There is also a secondary section of the dataset (\cref{sec:secondary}), which includes both calibration segments and other interesting and unique segments.
Each section highlights a specific subset of challenges in the multimotion estimation problem and acts as a scaffold for improving performance on increasingly complex scenes.

\begin{table}[t] 
	\caption{Sensors used in data collection.}
	\centering
	\begin{tabular}{llccc}
		\toprule
		Sensor Type & Model & Rate & Notes & \\
		\midrule
		Stereo & Bumblebee XB3 & 16 Hz & 1280$\times$960 & \\
		RGB-D & Realsense D435 & 30 Hz & 640$\times$480 & \\
		IMU & Microstrain 3DM-GX4-45 & 500 Hz &  & \\
		Pose & Vicon Motion Capture & 200 Hz &  & \\
	\end{tabular}
	\label{tab:sensors}
\end{table}

\subsection{Calibration} \label{sec:dataset:calibration}
The cameras and IMU were statically affixed to the collection platform, and their coordinate frames were manually aligned with z forward and y down, i.e., standard camera coordinate frames (\cref{fig:apparatus:diagram}).  
The Vicon motion capture system maintains a coordinate frame for each tracked object, including the sensor apparatus, but this frame is arbitrarily aligned to the sensor frames. 
The calibration between individual sensors and between the Vicon system and the sensor apparatus were each calculated from a sequence in the dataset.

The calibration between the sensors on the apparatus is calculated using Kalibr \cite{furgale2013}. The sensor apparatus is moved in front of a known, static calibration pattern, e.g., a checkerboard, such that all three cameras can observe the pattern and the IMU is sufficiently exercised in all directions.
Because the size and shape of the calibration pattern is known, the calibration problem becomes one of minimizing the geometric reprojection error between each camera.
To determine the calibration of the IMU, similar visual geometry techniques are employed to estimate the pose of the camera relative to the calibration pattern, and the camera-IMU calibration is determined by minimizing the difference between the poses of the two sensors.
Kalibr was used to determine the extrinsic calibration between the sensors, and the intrinsic calibrations it provides are also included in the dataset, though these are likely to be less accurate than those provided by the cameras themselves.

The calibration between the Vicon system and the sensor apparatus was calculated by aligning the visual odometry estimate of a test sequence with its Vicon ground truth.
To do this, the apparatus is moved through a static scene~and the trajectory in the camera frame is determined with VO.
Assuming the distribution of features and the trajectory estimator are unbiased, the calibration between the camera and Vicon frames is given by the \se[3]{} transform that minimizes the error between the VO trajectory and the Vicon trajectory. 
The extrinsic calibration between the left Bumblebee XB3 camera and the Vicon frame was calculated using a stereo VO pipeline, and is included in the dataset.\looseness=-1

The calibrations given are sufficient to determine the calibrations between any pair of sensors (\cref{fig:apparatus:diagram}). 
The data segments used to calculate the calibrations are included in the dataset for completeness (\cref{sec:secondary:calibration}).

\subsection{Data Format}
Each segment of the dataset consists of data from each of the sensors (\cref{tab:sensors}). 
This includes 1280x960 RGB image pairs from the Bumblebee XB3; 640x480 RGB and 16-bit raw and color-aligned depth images from the Realsense D435; 6-DOF orientation and acceleration information from the Microstrain IMU; and 6-DOF orientation and position information for each object, including the sensor apparatus, from the Vicon motion capture system. 
The folder name for each segment is of the form \texttt{<section>\_<\# object motions>\_<camera motion>}.\looseness=-1

\begin{table}[t] 
	\centering
	\caption{Primary data segments and their characteristics.}\label{tab:primary}
	\begin{tabular}{llcc}
		\toprule
		Section & \begin{tabular}[c]{@{}c@{}}Camera\\Motion\end{tabular} & \begin{tabular}[c]{@{}c@{}}\# Object\\Motions\end{tabular} & Duration \\
		\midrule
		Swinging & static & 4 & 6m 00s \\
		& translational & 4 & 3m 45s \\
		& unconstrained & 4 & 6m 00s \\
		Occlusion & static & 2 & 8m 00s \\
		& translational & 2 & 3m 00s \\
		& unconstrained & 2 & 6m 00s \\
		Toy cars & static & 3 & 3m 00s \\
		& static & 6 & 4m 45s \\
		& translational & 3 & 2m 00s \\
		& translational & 6 & 3m 00s \\
		& unconstrained & 3 & 6m 30s \\
		& unconstrained & 6 & 3m 00s \\
		& robot & 6 & 2m 00s 
	\end{tabular}
	\vspace{-1mm}
\end{table}

The data for each segment of the dataset is grouped by sensor.
The stereo images from the Bumblebee XB3 have been debayered, undistorted, and rectified using the manufacturer's calibration, and they are provided as individual, indexed left and right image frames in \texttt{stereo.tgz}.
A CSV file provides the necessary map from frame index to timestamp for each synchronized pair of images.
The color and aligned depth images from the Realsense D435 are synchronized and provided as individual images in \texttt{rgbd.tgz} with their own CSV lookup.
The raw depth images are not well-synchronized, so they are provided in a separate \texttt{raw\_depth.tgz} with its own CSV file.
The camera intrinsics and extrinsics for the sensor platform as calculated by Kalibr are included as a \texttt{yaml} file. 
The manufacturer-provided calibrations are given in a similar format where available.
The \se[3]{} calibration transform from the left Bumblebee XB3 camera and Vicon object frame for the sensor apparatus is given in the same \texttt{yaml} format.

The rotation quaternion, angular velocity, and linear acceleration information from the IMU is provided as timestamped rows of a CSV file.
Likewise, the pose information from the Vicon is provided as quaternions and translation vectors for each object in timestamped rows of a CSV file.
The \se[3]{} passive/alias transformation from the `world' frame to the object frame at each time step is also included in each row.

\begin{figure}[t]	
	\centering
	\subfloat[\label{fig:swinging:image}]{%
		\includegraphics[width=\columnwidth]{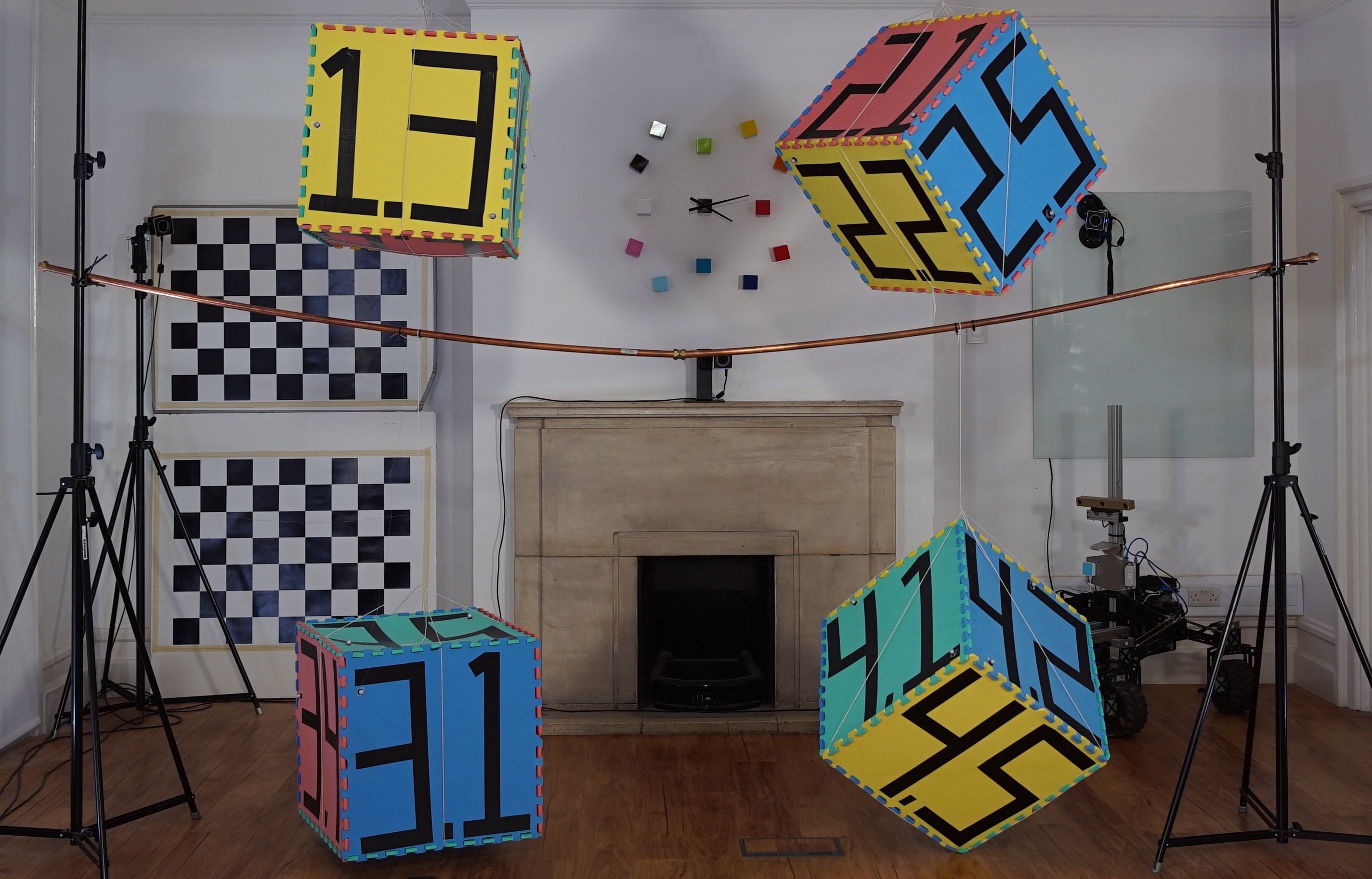}%
	}
	
	\subfloat[\label{fig:swinging:diagram}]{%
		\includegraphics[width=.96\columnwidth]{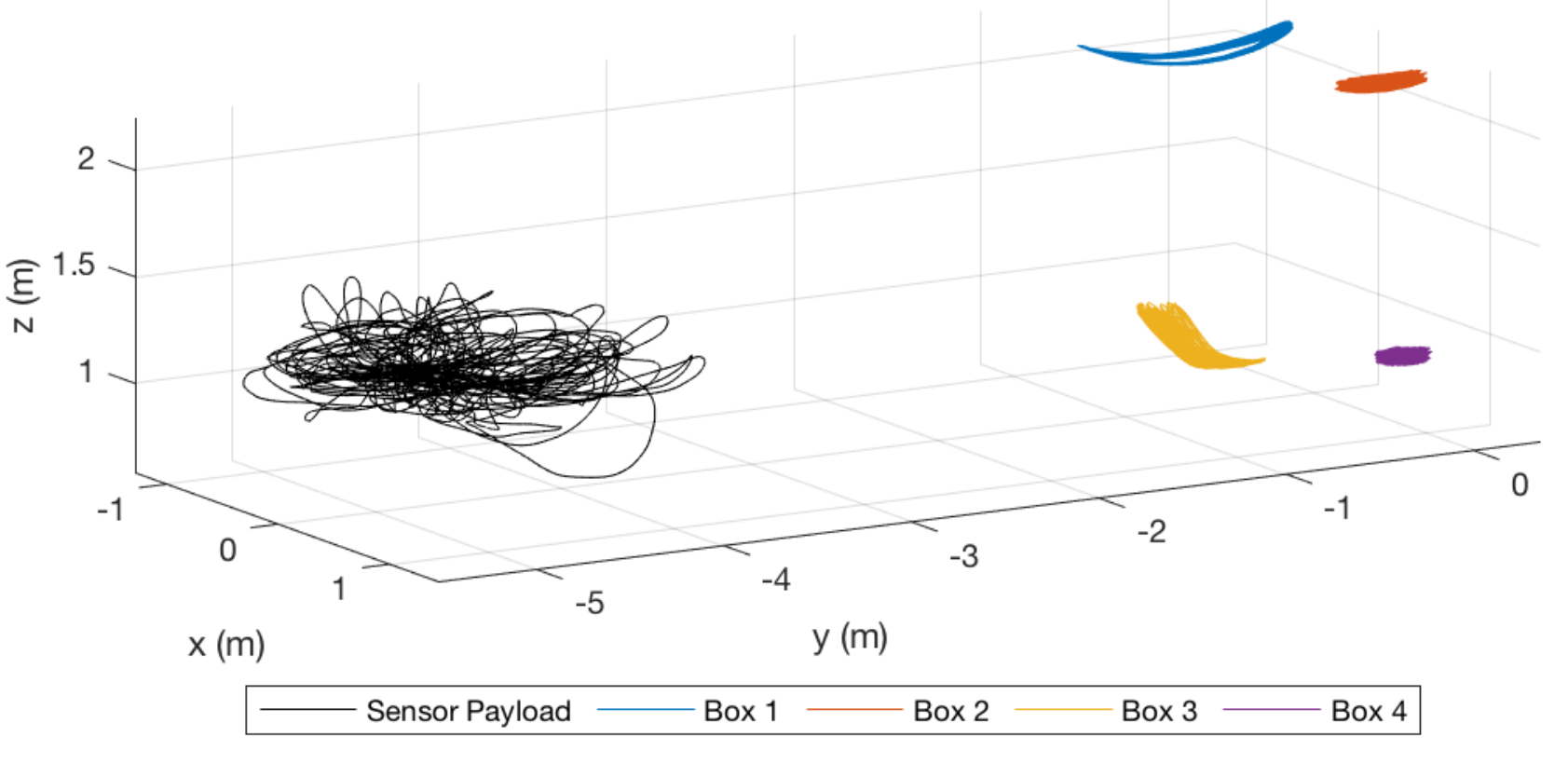}%
	}%
	\caption{In the \textit{Swinging Blocks} portion of the dataset, the cameras (black) observe four independently swinging blocks. The blocks swing and rotate as pendulums to generate four distinct \se[3]{} motions. This segment is a basic multimotion scene requiring full \se[3]{} motion estimation of multiple bodies without complications such as occlusions or collisions. The scene is shown in (a) and the Vicon ground-truth motion data from the unconstrained \se[3]{} segment is shown in (b).} \label{fig:swinging}
\end{figure}
The raw ROS \texttt{bag} file for each individual sensor stream is also included with each segment for those who wish to stream the data into ROS-based systems.

\subsection{Primary Data Segments} \label{sec:dataset:primary}
The dataset contains three primary data segments that form a foundation for developing and evaluating multimotion estimation techniques (\cref{tab:primary}).
The first segment consists of four independently moving blocks and acts as an example of \se[3]{} multimotion estimation.
The second segment also features moving blocks but involves significant occlusion and is designed to test the ability of an approach to handle missing observations.
The final segment includes multiple radio-controlled cars that present many different motions and sources of occlusion.
This final segment is intended to represent the type of multimotion estimation problem we believe is necessary to solve in robotics.

\begin{figure}[t]	
	\centering
	\subfloat[\label{fig:occlusion:image}]{%
		\includegraphics[width=\columnwidth]{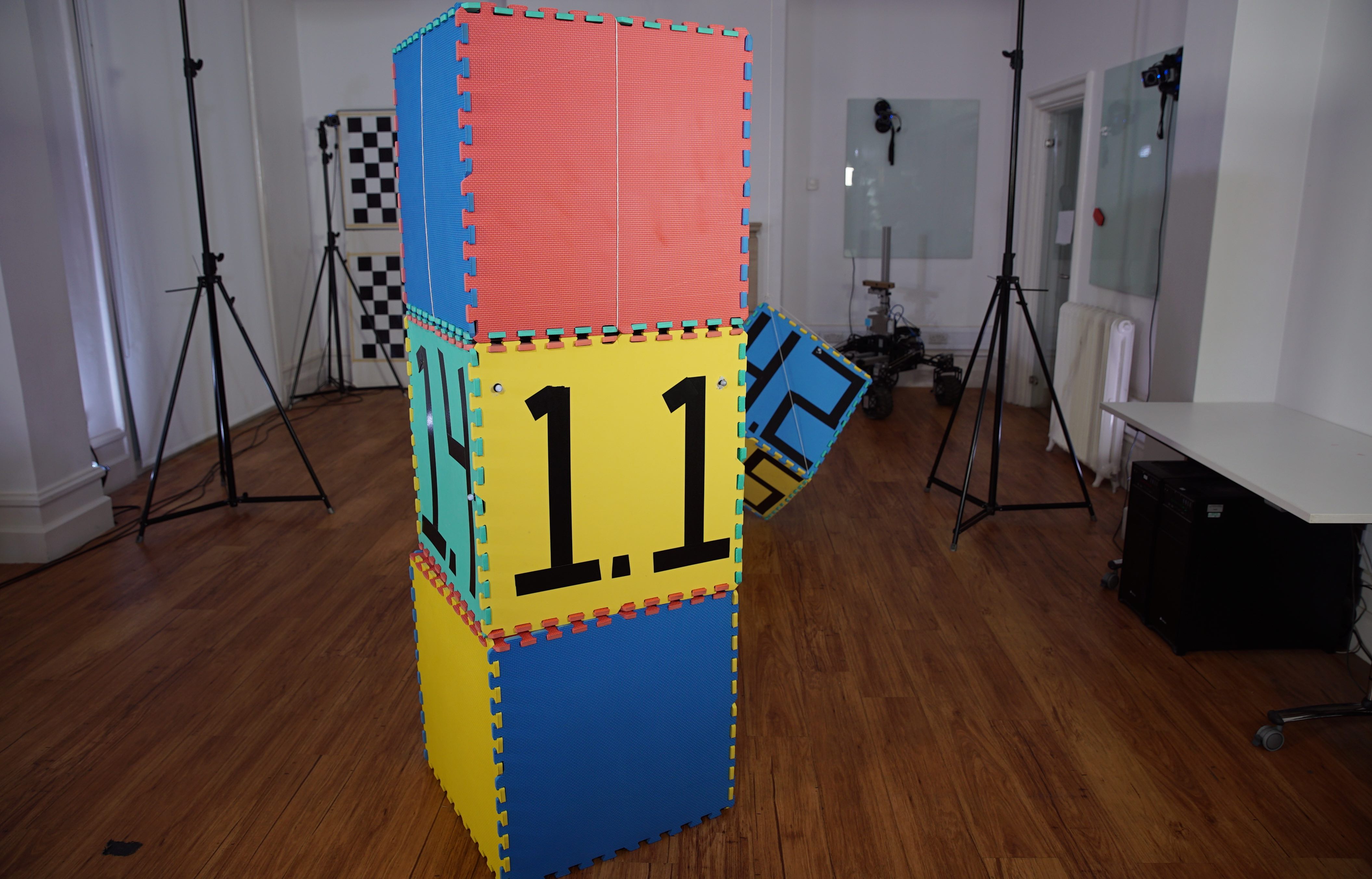}%
	}
	
	\subfloat[\label{fig:occlusion:diagram}]{%
		\includegraphics[width=\columnwidth]{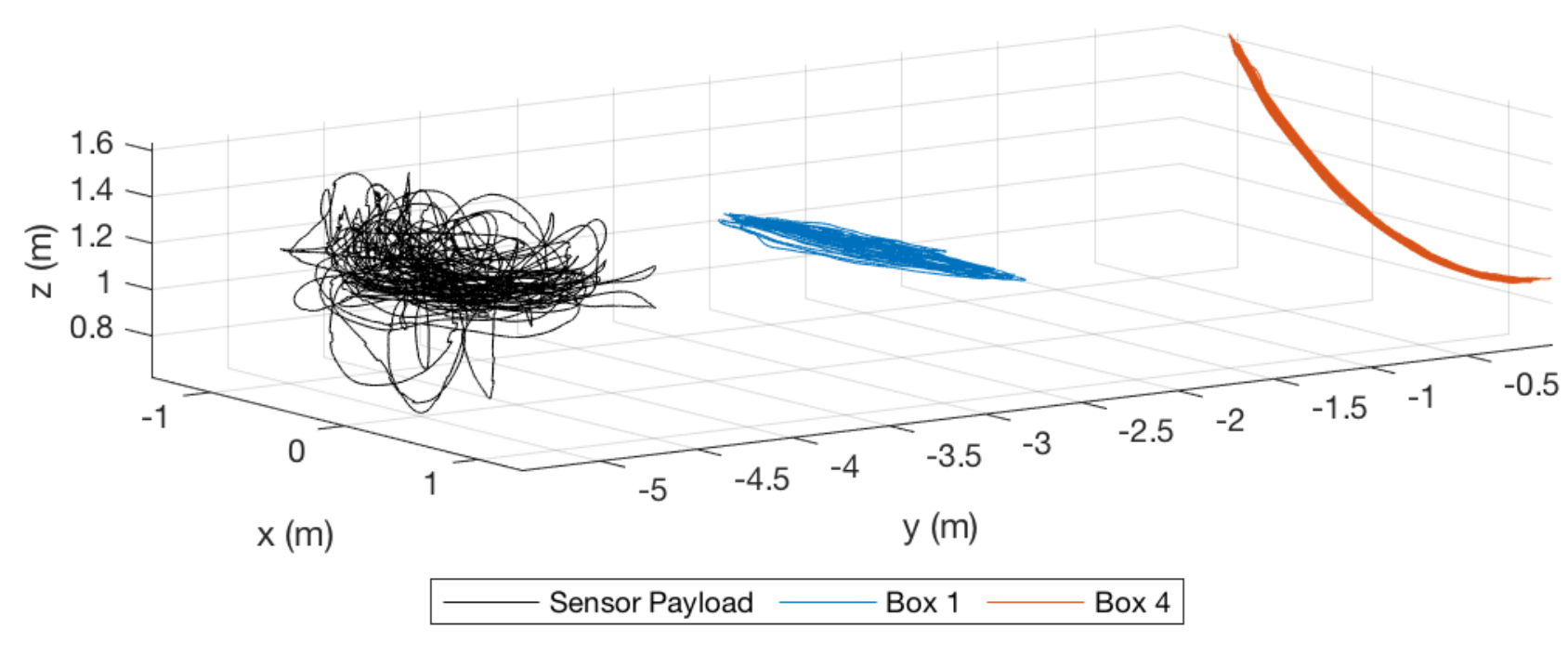}%
	}%
	\caption{The \textit{Occluding Blocks} portion of the dataset introduces both partial and total occlusions to the multimotion estimation problem. A single swinging block (red) is observed by the sensor (black) while a sliding block tower (blue) intermittently occludes it. The motion of both blocks creates partial occlusions, where the swinging block can still be partially observed, and total occlusions, where estimations must be extrapolated in the absence of direct observations. This is complicated further by the fact that the occluded block occasionally changes direction while it is occluded. Successfully estimating these motions and interpolating through occlusions is integral to fully addressing the multimotion estimation problem. The scene is shown in (a) and the Vicon ground-truth motion data from the unconstrained \se[3]{} segment is shown in (b).\looseness=-1} \label{fig:occlusion}
\end{figure}

\subsubsection{Swinging Blocks} \label{sec:dataset:swinging}

The swinging blocks dataset consists of four independently moving cubes (\cref{fig:swinging}). 
Each block has a different pendular motion and is observed without occlusion by the camera in static and dynamic segments.

Each block swings with an independent, repetitive \se[3]{} motion and remains spatially separated in both image- and world-space. 
These segments are designed to isolate the challenge of accurately segmenting and estimating the motions of multiple objects without complications caused by occlusions.

\subsubsection{Occluding Blocks} \label{sec:dataset:occlusion}
The occlusion segments involve a swinging block that is partially and fully occluded by a sliding tower of blocks (\cref{fig:occlusion}). 
The motion of the swinging block is similar in complexity to those of the previous segments, and the sliding block tower moves in \se[2]{} in the foreground. 
This presents a simplified occlusion example in the multimotion estimation problem. 

\begin{figure}[t]	
	\centering
	\subfloat[\label{fig:cars:image}]{%
		\includegraphics[width=\columnwidth]{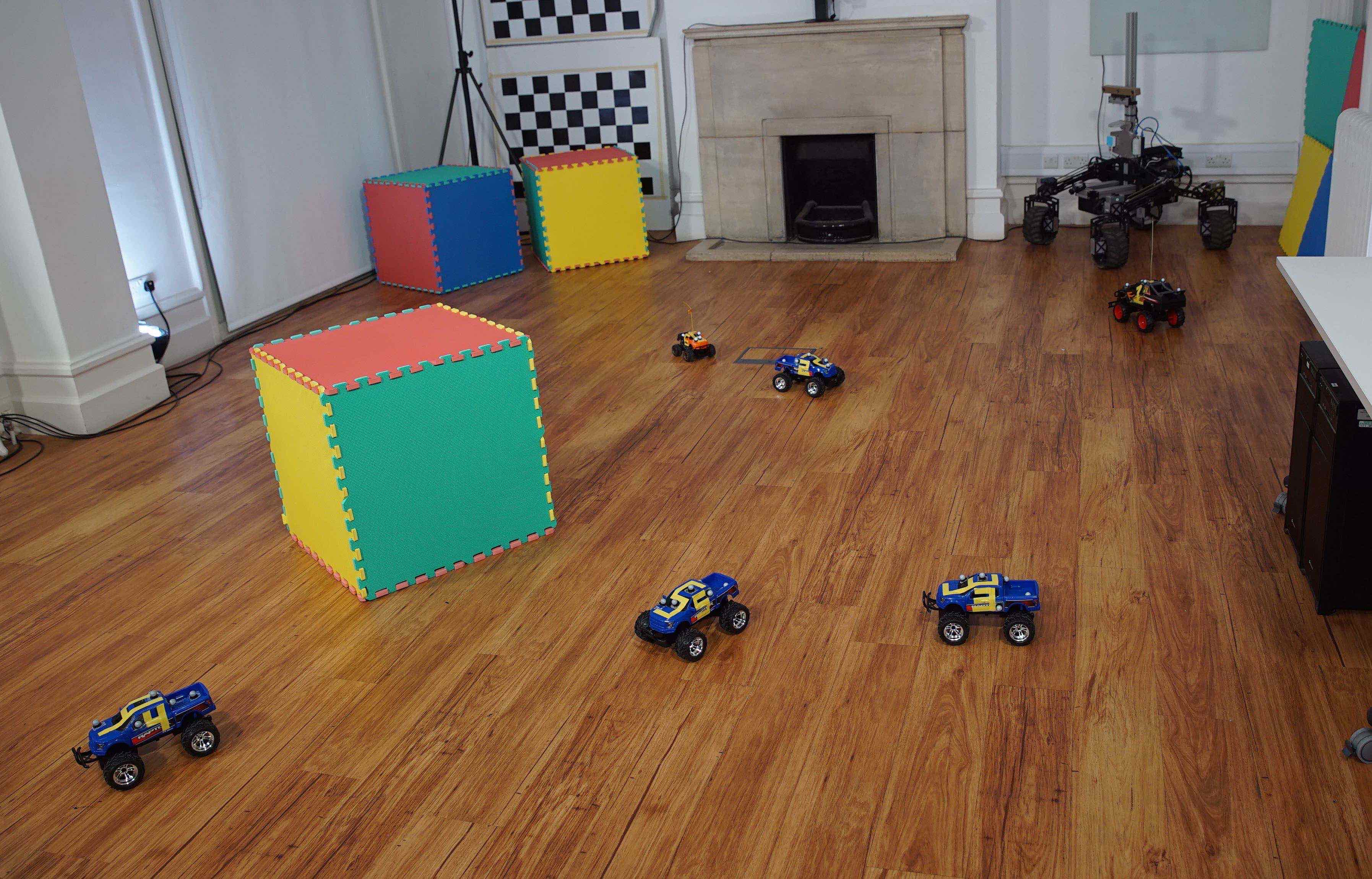}%
	}\vspace{-2mm}
	\subfloat[\label{fig:cars:diagram}]{%
		\includegraphics[width=\columnwidth]{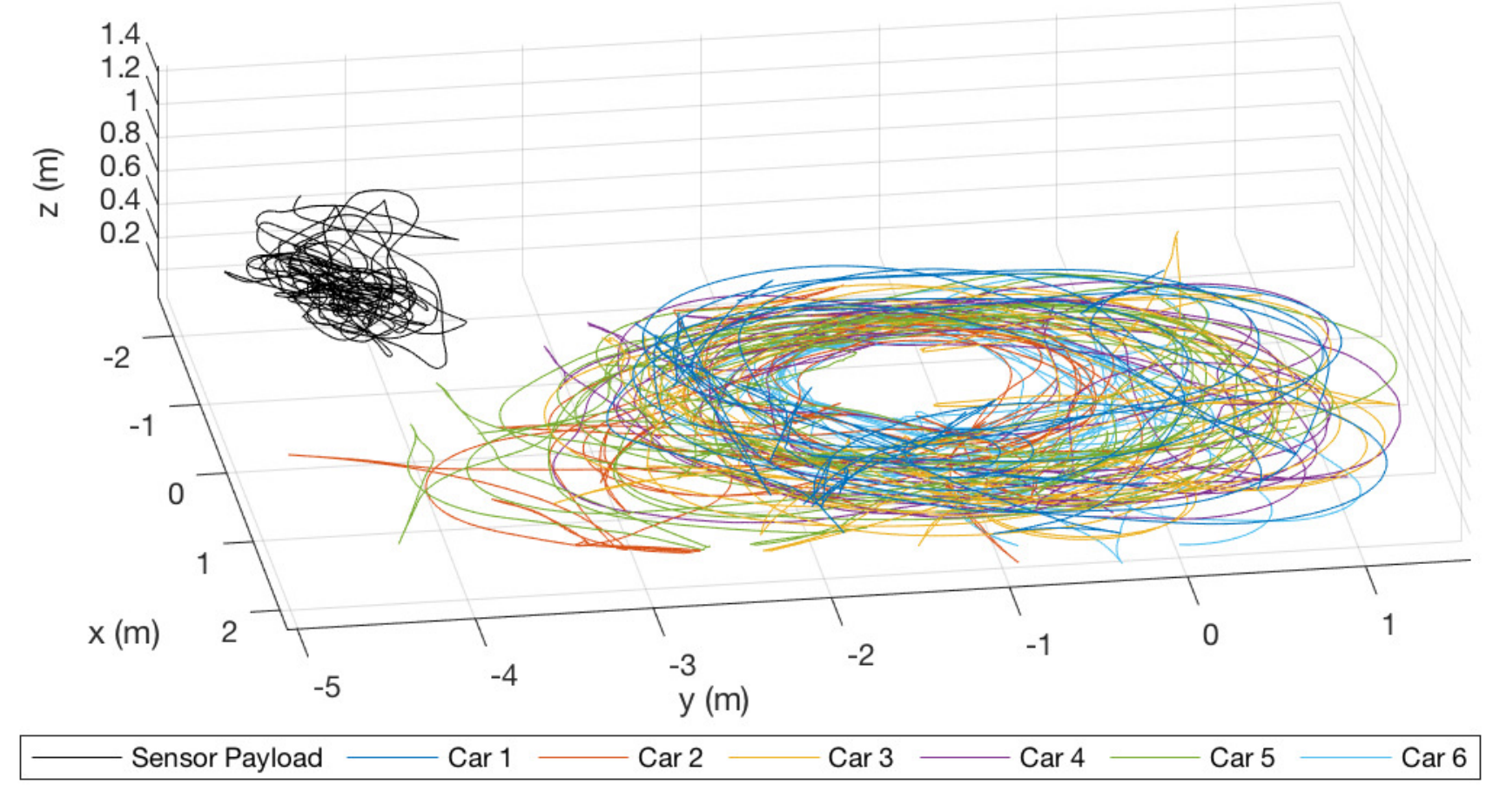}%
	}%
	\caption{The \textit{Toy Cars} segments of the dataset are the most complex and involve the highest number of independent motions. The cameras (black) observe up to six individual radio-controlled toy cars as they maneuver around a block that also adds occlusions. The motion of each car is restricted to \se[2]{}, but the dynamic cameras move in $\mathbb{R}^{3}$ and \se[3]{}. This section also includes a robot-mounted segment (\cref{fig:marquee}), where perspective effects are significantly increased and occlusions caused by the dynamic objects themselves are more frequent. The scene is shown in (a) and the Vicon ground-truth motion data from the unconstrained \se[3]{} segment is shown in (b).}  
	\label{fig:cars}
\end{figure}

\begin{figure}[t]	
	\centering
	\subfloat[\label{fig:pinwheel:photo}]{%
		\includegraphics[width=\columnwidth]{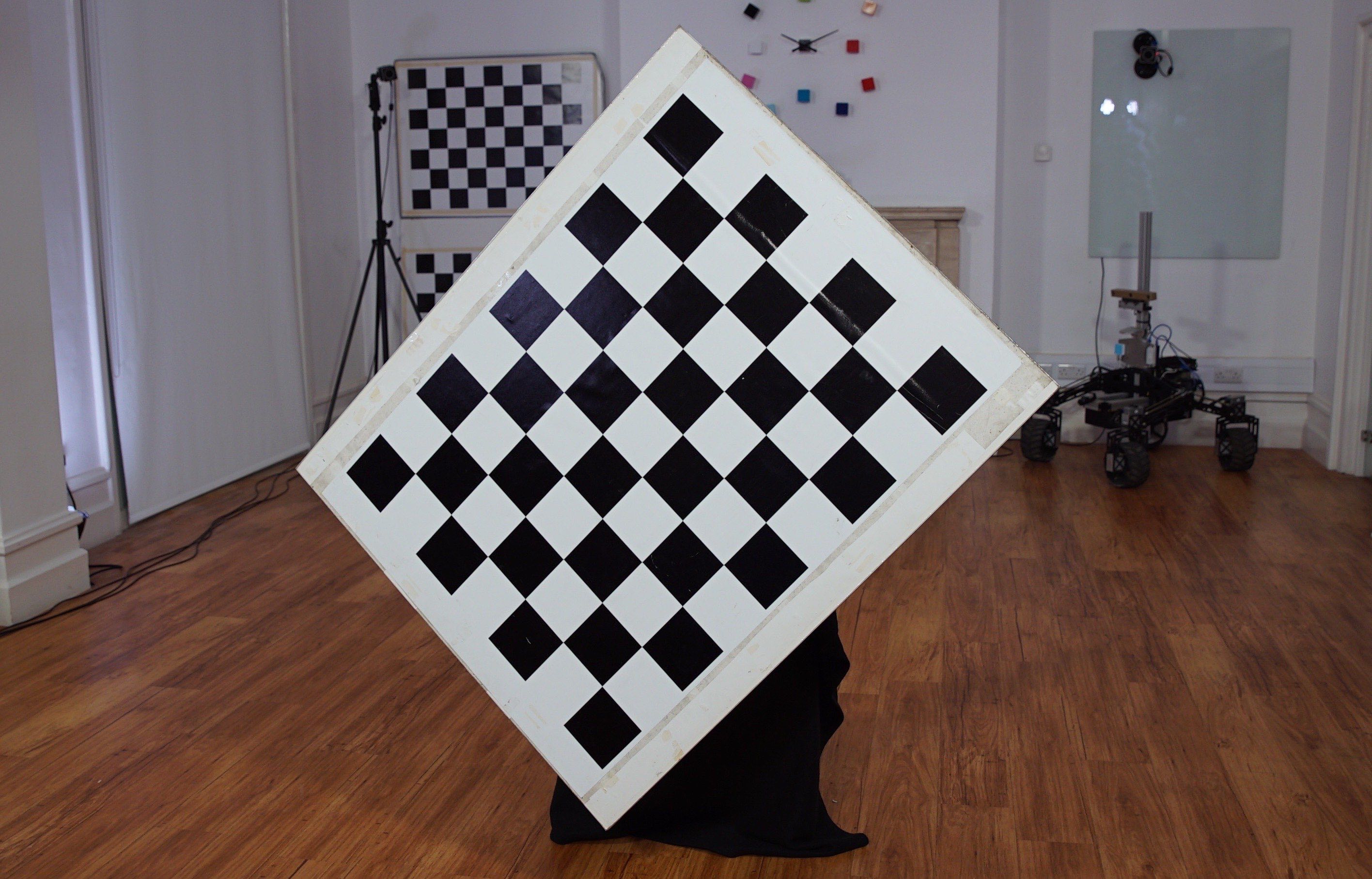}%
	}%
	
	\subfloat[\label{fig:pinwheel:diagram}]{%
		\includegraphics[clip,width=\columnwidth,page=1]{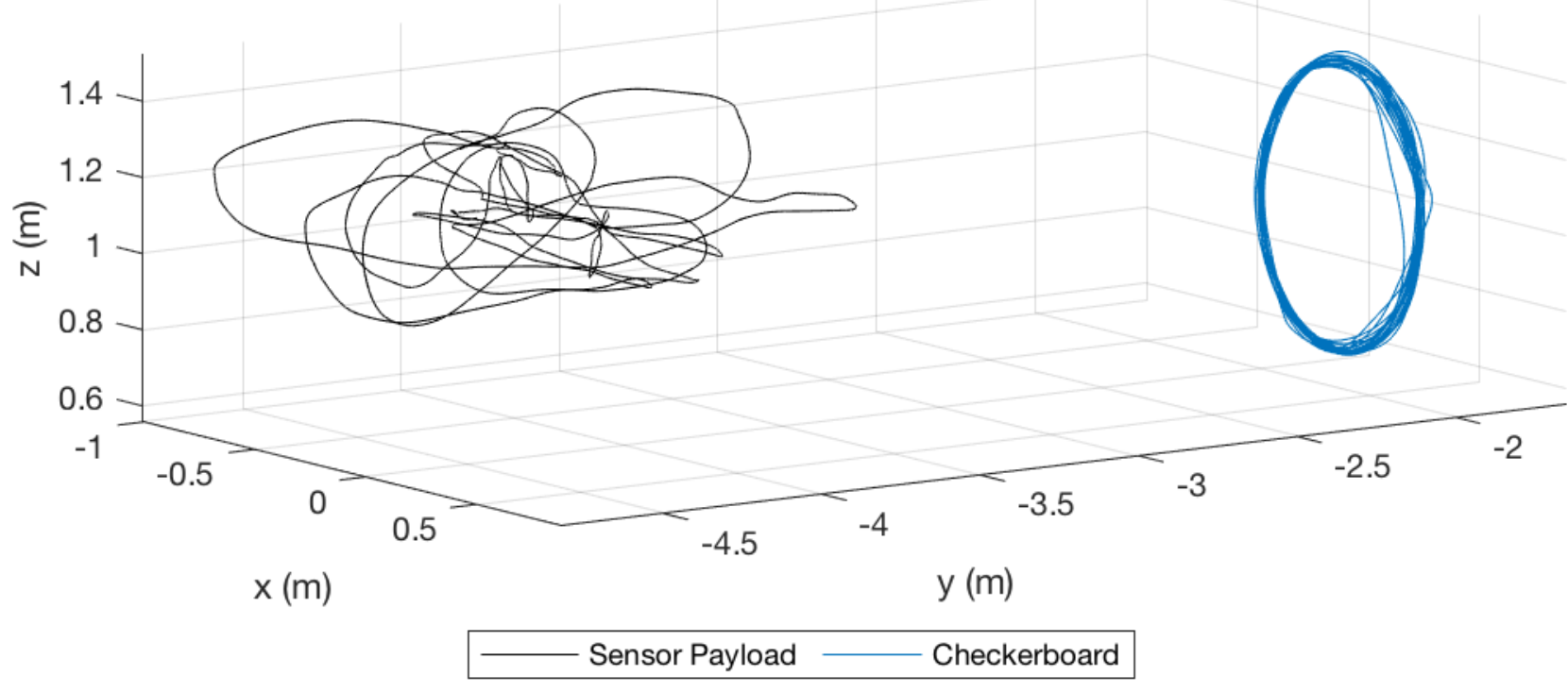}%
	}%
	\caption{The unconstrained pinwheel segment of the dataset consists of a rotating checkerboard pattern (blue), observed by a dynamic sensor (black). The board rotates primarily in the image plane of the camera, changing rotation direction halfway through each segment. This segment tests the ability of a multimotion estimation technique to estimate rotations about the optical axis of the camera. The checkerboard and scene are shown in (a) and the Vicon ground-truth motion data from the unconstrained \se[3]{} segment is shown in (b).\looseness=-1} 
	\label{fig:pinwheel}
\end{figure}

The loss of information caused by occlusion poses several additional challenges to the multimotion estimation problem.
Partial occlusions change the observed shape of the occluded object, which may be challenging for many appearance-based tracking techniques.
In these sequences, the occluded and occluding objects are made of visually similar blocks which poses an even greater challenge for appearance-based approaches.\looseness=-1

Full occlusions are even more challenging due to the lack of direct observations. 
The lack of direct observations means estimations must be extrapolated forward in time. 
If the occluded body can be recognized when it becomes unoccluded, these extrapolated estimations can be interpolated to more intelligently explain the unobserved motion. 
Na\"ive linear interpolation schemes can reasonably estimate occluded motions that are largely constant, but accurately estimating more complex occluded motion requires further research.

\subsubsection{Toy Cars} \label{sec:dataset:cars}
The final data segments involve a varying number of RC cars moving around the room (\cref{fig:cars}). 
A single block provides an obstacle for the cars and a source of occlusion.
The complexity of the segments is varied according to both the number of object motions and the motion of the camera, either static, $\mathbb{R}^{3}$, or \se[3]{} as in the previous segments.

Although the car motions are restricted to \se[2]{}, these segments are the still challenging to estimate. 
The cars are significantly smaller in size than the objects used in the other segments, which poses a greater challenge in segmenting and estimating the motions. 
Likewise, four of the cars are of the same model, which also challenges appearance-based techniques.\looseness=-1

\begin{table}[t!] 
	\centering
	\caption{Secondary data segments and their characteristics.}\label{tab:secondary}
	\begin{tabular}{llcc}
		\toprule
		Section & \begin{tabular}[c]{@{}c@{}}Camera\\Motion\end{tabular} & \begin{tabular}[c]{@{}c@{}}\# Object\\Motions\end{tabular} & Duration \\
		\midrule
		calibration & vicon &  & 3m 15s \\
		& extrinsic &  & 2m 45s \\
		& vicon\_no\_imu\footnotemark[1]{} &  & 2m 50s \\
		& extrinsic\_no\_imu\footnotemark[1]{} &  & 5m 45s \\
		Pinwheel & static & 1 & 1m 05s \\
		& unconstrained & 1 & 1m 15s \\
		Fixed Occlusion & static & 1 & 3m 25s  \\
		& translational & 1 & 1m 50s \\
		& unconstrained & 1 & 4m 50s \\
		Toy cars & unconstrained & 1 & 3m 00s \\
		& static\footnotemark[1]{} & 2 & 2m 00s \\
		& static\footnotemark[1]{} & 6 & \hspace{-1.4mm}10m 00s \\
		& unconstrained\footnotemark[1]{} & 6 & 5m 05s \\
		& robot\footnotemark[1]{} & 6 & 6m 50s 
	\end{tabular} 
\end{table}

An additional toy car segment was recorded with the sensor apparatus mounted on a Clearpath Robotics Jackal (\cref{fig:marquee}). 
Recording from the perspective of a robotic vehicle reduces the complexity of the sensor egomotion from \se[3]{} to \se[2]{}, but also greatly increases the perspective effects for the cameras. 
The broad depth of field in this scene also highlights problems with the affine camera model. 
\footnotetext[1]{These files have no IMU data and use a different calibration.}
The model relies on an orthogonal projection from world- to image-space, which leads to significant errors in scenes with large depth disparity or field of view.
This means the Jackal segment may be particularly challenging for techniques that rely on the affine camera model.

\subsection{Secondary Data Segments} \label{sec:secondary}
Several additional data segments are included in the dataset either for completeness or because they exhibit particularly interesting motions (\cref{tab:secondary}). 

\subsubsection{Calibration} \label{sec:secondary:calibration}
The data used to determine the calibration between individual sensors and from the sensor apparatus to the vicon system is provided for completeness.
The first calibration segment observes a static 7x9 checkerboard pattern (6x8 internal corners) with $0.1$m-long squares while the sensor platform is moved through a repetitive progression of prescriptive motions: tilt (pitch), pan (yaw), roll, truck (x), pedestal (z), and dolly (y). 
This progression provides the necessary motions for finding the extrinsic calibrations of the sensor apparatus.\looseness=-1

The second calibration segment involves the sensor apparatus moving around the room in three large circles. 
The motion is continuously clockwise so that, given a consistently biased estimator, the odometry drift will increase monotonically.
This means a full-batch minimization of the VO trajectory error can be used to calibrate the camera and Vicon frames.

\begin{figure}[t]
	\centering
	\includegraphics[clip,width=\columnwidth,page=1]{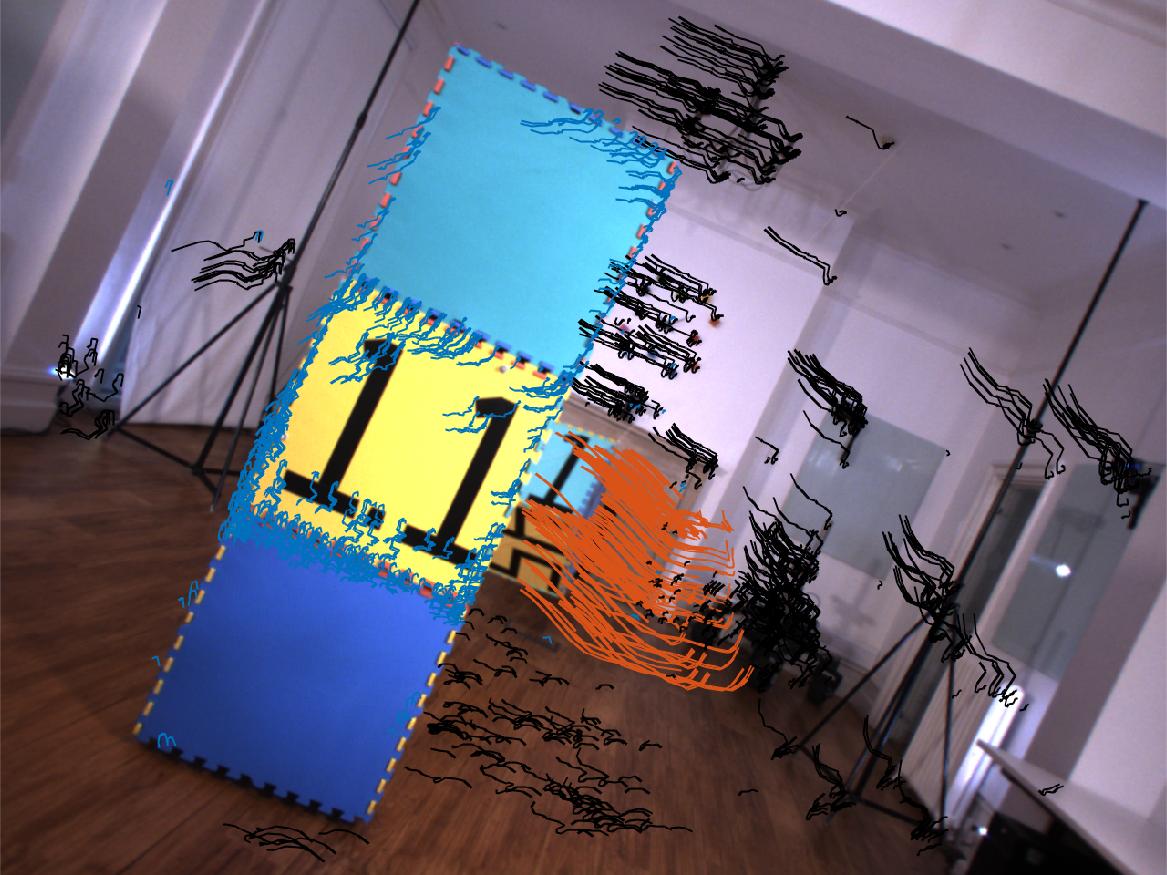}
	\caption{Motion segmentation produced by Multimotion Visual Odometry \cite{judd2018} for the unconstrained $SE\left(3\right)$ camera segment of the occlusion section of the dataset. The egomotion of the camera is estimated from the static points in the scene shown in black. The occluded block (red) is tracked until it moves behind the block tower (blue).}
	\label{fig:segmentation}
\end{figure}

\subsubsection{Pinwheel} \label{sec:dataset:pinwheel}
The simplest piece of the dataset consists of a single rotating checkerboard observed with static and  unconstrained $SE\left(3\right)$ camera motions.
This simple, largely rotational motion highlights limitations in many flow-based and tracking techniques, which struggle to estimate rotations about the optical axis. 
This type of motion is less common in existing datasets, such as those focused on driving scenes, and it is therefore important to include in this dataset to ensure multimotion estimation techniques can accurately estimate it.

This segment can also be challenging for feature-based techniques. 
The checkerboard pattern provides strong corners for feature detectors, but the repetitive texture can lead to aliasing when there is significant object and/or camera motion.

\begin{figure}[t]
	\centering
	\subfloat[Camera Egomotion]{
		\centering
		\includegraphics[clip,width=.98\columnwidth,page=1]{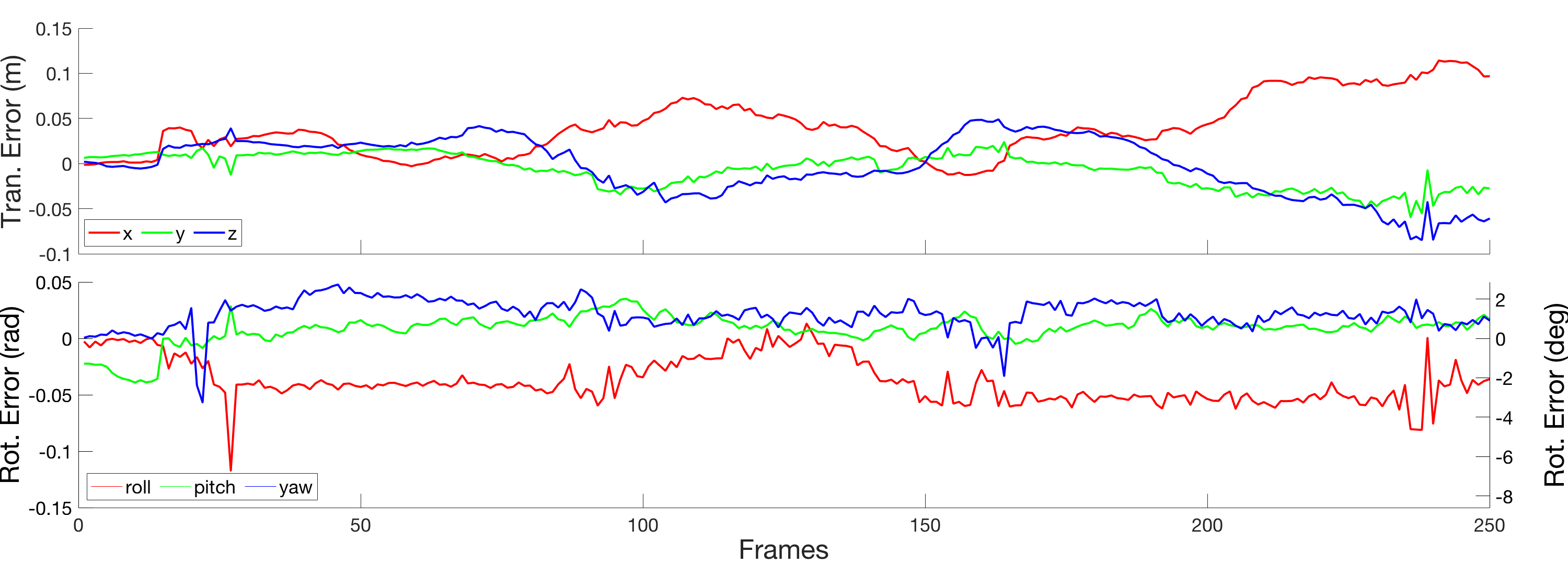}%
	}
	
	\subfloat[Box 1 (Occluding Tower)]{
		\centering
		\includegraphics[clip,width=.98\columnwidth,page=1]{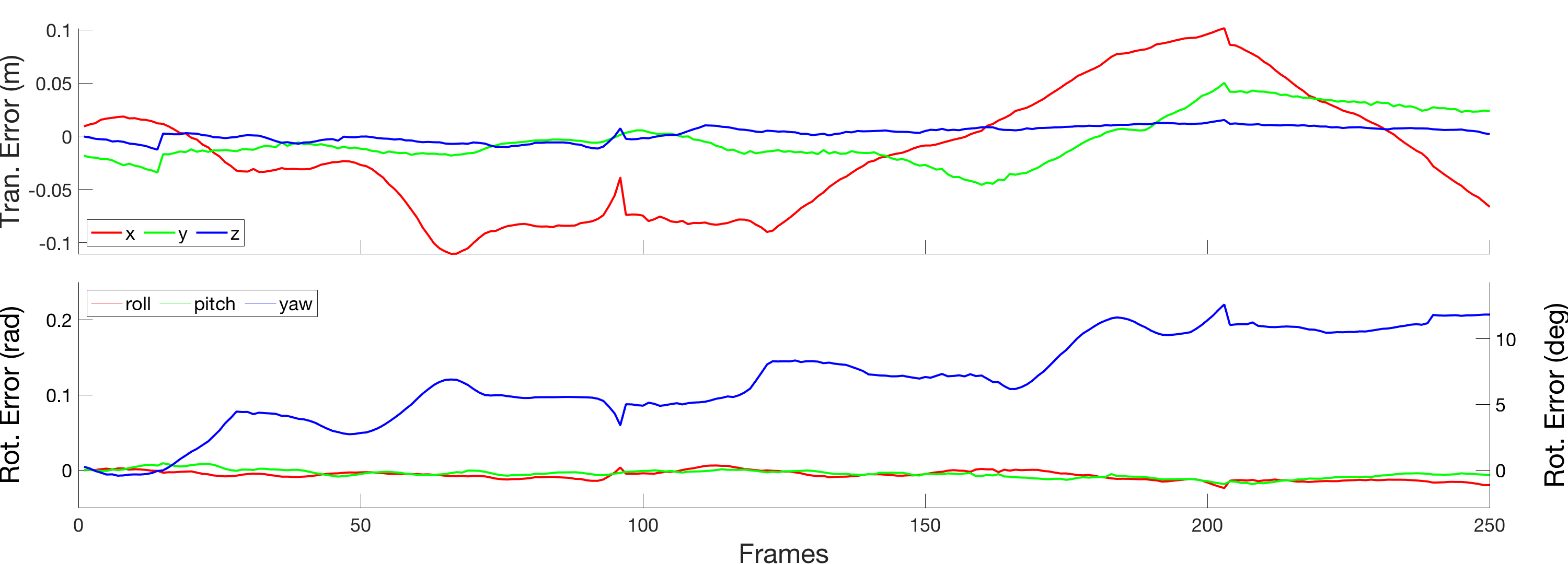}%
	}
	
	\subfloat[Box 4 (Occluded Box)]{
		\centering
		\includegraphics[clip,width=.98\columnwidth,page=1]{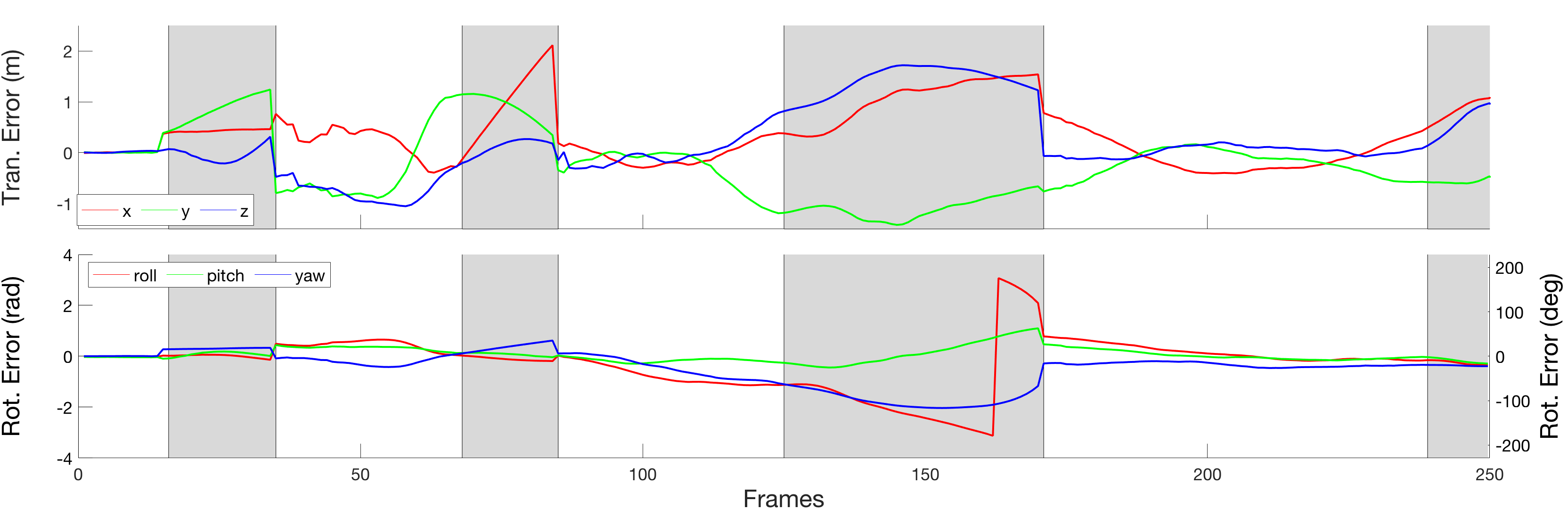}%
	}
	\caption{The translational and rotational errors for the estimated motion of each object over the length of the estimation window as compared to ground-truth trajectory data. Errors are reported in an arbitrary geocentric frame with the z-axis up, and arbitrary x- and y-axes. The swinging block (c) was occasionally occluded by the block tower (b) and those periods are shown in grey. These occlusions result in errors in the trajectory estimate as it is na\"ively extrapolated forward in time. During the first two occlusions, the block continues in the same direction while unobserved, so the errors grow linearly. The block changes direction during the third occlusion, which greatly increases the error, especially in rotation. When the block becomes unoccluded, the errors decrease significantly as direct estimation resumes. \looseness=-1}
	\label{fig:results}
\end{figure}

\subsubsection{Additional Segments}
Eight additional segments are included to provide more accessible or varied examples of the primary data segments.
The first three segments, \texttt{fixed\_occlusion\_*}, are identical to the \textit{Occluding Blocks} segments except the occluding tower remains static.
This poses a simpler motion estimation problem involving occlusion and contributes to the scaffolding design of the dataset.

The next two segments, \texttt{cars\_1\_unconstrained} and \texttt{cars\_2\_static\_no\_imu}, are simplifications of the primary \textit{Toy Cars} segments. 
In \texttt{cars\_1\_unconstrained}, a single radio-controlled car is observed with dynamic $SE\left(3\right)$ sensor motion.
Likewise, the \texttt{cars\_2\_static\_no\_imu} presents an example of two cars observed by the static sensor apparatus, but does not have IMU data due to sensor failure.
These segments provide simpler examples of the \textit{Toy Cars} scene which can be used to progress to the primary segments.

The final three segments, \texttt{cars\_6\_}*\texttt{\_no\_imu}, are longer data segments, identical to their respective \textit{Toy Car} segments but with a different calibration and without IMU due to sensor failure.
These segments are included for those who seek longer experiments but do not require IMU data.

\section{Results} \label{sec:experiments}
The utility of this dataset is demonstrated by using MVO \cite{judd2018} to estimate the motions in the unconstrained $SE\left(3\right)$ occlusion segment (\cref{fig:segmentation,fig:results}).
The performance of MVO illustrates the use of a simple extrapolation scheme to estimate occluded~objects.

When an occlusion is detected, the last estimate for the occluded motion is na\"ively extrapolated forward in time using a constant velocity assumption. 
This interpolation scheme is adequate for situations in which the motion of the occluded object is constant, but this segment also includes situations where the object changes direction during occlusions.
To improve performance, the trajectory estimate can be retrospectively interpolated through the occlusion when the object is detected again by using linear interpolation in $\mathfrak{se}\left(3\right)$ algebra space.
More accurate motion priors \cite{barfoot2015, tang2018} may be able to better estimate this more complicated motion.
This shows there is significant room for further work in estimating motion through occlusion, and this dataset includes a useful gradient of difficulties to measure that progress. 

\section{Conclusion} \label{sec:conclusion}
This paper introduces the Oxford Multimotion Dataset, a collection of image sequences designed to foster development of multimotion estimation algorithms. 
Existing datasets have proven to be useful benchmarks in areas such as egomotion estimation and motion segmentation, but none are fully suitable for multimotion estimation. 
The Oxford Multimotion Dataset includes three primary scenes observed from a sensor apparatus that includes a stereo camera, an RGB-D camera, and an IMU.
Most importantly, the dataset also includes ground-truth trajectory information for \emph{every} motion in each scene collected from a Vicon motion capture system.

The dataset is designed as a scaffold to foster the development and testing of new metric multimotion estimation techniques. 
Each section highlights particular aspects of the problem and presents unique challenges, culminating in a complex toy car segment representative of many real-world scenarios.\looseness=-1

The data segments and associated scripts can be downloaded from \dataurl.

\section*{Acknowledgment}
We would like to thank the members of the Oxford Robotics Institute and Somerville College who contributed their driving skills to the collection of this dataset.


{\small\begin{thebibliography}{10}
		\providecommand{\url}[1]{#1}
		\csname url@samestyle\endcsname
		\providecommand{\newblock}{\relax}
		\providecommand{\bibinfo}[2]{#2}
		\providecommand{\BIBentrySTDinterwordspacing}{\spaceskip=0pt\relax}
		\providecommand{\BIBentryALTinterwordstretchfactor}{4}
		\providecommand{\BIBentryALTinterwordspacing}{\spaceskip=\fontdimen2\font plus
			\BIBentryALTinterwordstretchfactor\fontdimen3\font minus
			\fontdimen4\font\relax}
		\providecommand{\BIBforeignlanguage}[2]{{%
				\expandafter\ifx\csname l@#1\endcsname\relax
				\typeout{** WARNING: IEEEtran.bst: No hyphenation pattern has been}%
				\typeout{** loaded for the language `#1'. Using the pattern for}%
				\typeout{** the default language instead.}%
				\else
				\language=\csname l@#1\endcsname
				\fi
				#2}}
		\providecommand{\BIBdecl}{\relax}
		\BIBdecl
		
		\bibitem{sturm2012}
		J.~Sturm, N.~Engelhard, F.~Endres, W.~Burgard, and D.~Cremers, ``A benchmark
		for the evaluation of {RGB-D} {SLAM} systems,'' in \emph{IROS}, Oct. 2012.
		
		\bibitem{geiger2012}
		A.~Geiger, P.~Lenz, and R.~Urtasun, ``Are we ready for autonomous driving? the
		{KITTI} vision benchmark suite,'' in \emph{CVPR}, 2012, pp. 3354--3361.
		
		\bibitem{tron2007}
		R.~Tron and R.~Vidal, ``A benchmark for the comparison of 3-{D} motion
		segmentation algorithms,'' in \emph{CVPR}, 2007, pp. 1--8.
		
		\bibitem{ochs2014}
		P.~Ochs, J.~Malik, and T.~Brox, ``Segmentation of moving objects by long term
		video analysis,'' \emph{PAMI}, vol.~36, no.~6, pp. 1187--1200, June 2014.
		
		\bibitem{gaidon2016}
		A.~Gaidon, Q.~Wang, Y.~Cabon, and E.~Vig, ``Virtual worlds as proxy for
		multi-object tracking analysis,'' in \emph{CVPR}, June 2016, pp. 4340--4349.
		
		\bibitem{dosovitskiy2017}
		A.~Dosovitskiy, G.~Ros, F.~Codevilla, A.~Lopez, and V.~Koltun, ``Carla: An open
		urban driving simulator,'' 2017, arXiv: 1711.03938 [cs.CV].
		
		\bibitem{judd2018}
		K.~M. Judd, J.~D. Gammell, and P.~Newman, ``{Multimotion Visual Odometry}
		({MVO}): Simultaneous estimation of camera and third-party motions,'' in
		\emph{IROS}, Madrid, Spain, 1--5 Oct. 2018.
		
		\bibitem{horn1981}
		B.~K. Horn and B.~G. Schunck, ``Determining optical flow,'' \emph{Artificial
			intelligence}, vol.~17, no. 1-3, pp. 185--203, 1981.
		
		\bibitem{vedula1999}
		S.~Vedula, S.~Baker, P.~Rander, R.~Collins, and T.~Kanade, ``Three-dimensional
		scene flow,'' in \emph{ICCV}, vol.~2, 1999, pp. 722--729.
		
		\bibitem{lenz2011}
		P.~Lenz, J.~Ziegler, A.~Geiger, and M.~Roser, ``Sparse scene flow segmentation
		for moving object detection in urban environments,'' in \emph{IV}, 2011, pp.
		926--932.
		
		\bibitem{menze2015}
		M.~Menze and A.~Geiger, ``Object scene flow for autonomous vehicles,'' in
		\emph{CVPR}, 2015, pp. 3061--3070.
		
		\bibitem{milan2016}
		A.~Milan, L.~Leal{-}Taix{\'{e}}, I.~D. Reid, S.~Roth, and K.~Schindler,
		``{MOT16:} {A} benchmark for multi-object tracking,'' 2016, arXiv: 1603.00831
		[cs.CV].
		
		\bibitem{kalal2012}
		Z.~Kalal, K.~Mikolajczyk, J.~Matas \emph{et~al.},
		``Tracking-learning-detection,'' \emph{PAMI}, vol.~34, no.~7, p. 1409, 2012.
		
		\bibitem{hartley2003}
		R.~Hartley and A.~Zisserman, \emph{Multiple view geometry in computer
			vision}.\hskip 1em plus 0.5em minus 0.4em\relax Cambridge university press,
		2003.
		
		\bibitem{costeira1998}
		J.~P. Costeira and T.~Kanade, ``A multibody factorization method for
		independently moving objects,'' \emph{IJCV}, vol.~29, no.~3, pp. 159--179,
		1998.
		
		\bibitem{vidal2004}
		R.~Vidal and R.~Hartley, ``Motion segmentation with missing data using power
		factorization and {GPCA},'' in \emph{CVPR}, 2004, pp. 310--316.
		
		\bibitem{li2007}
		T.~Li, V.~Kallem, D.~Singaraju, and R.~Vidal, ``Projective factorization of
		multiple rigid-body motions,'' in \emph{CVPR}, 2007, pp. 1--6.
		
		\bibitem{vidal2003}
		R.~Vidal and S.~Sastry, ``Optimal segmentation of dynamic scenes from two
		perspective views,'' in \emph{CVPR}, vol.~2, 2003, pp. 281--286.
		
		\bibitem{fischler1981}
		M.~A. Fischler and R.~C. Bolles, ``Random sample consensus: a paradigm for
		model fitting with applications to image analysis and automated
		cartography,'' \emph{Communications of the ACM}, vol.~24, no.~6, pp.
		381--395, 1981.
		
		\bibitem{schindler2006}
		K.~Schindler, U.~James, and H.~Wang, ``Perspective n-view multibody
		structure-and-motion through model selection,'' in \emph{ECCV}.\hskip 1em
		plus 0.5em minus 0.4em\relax Springer, 2006, pp. 606--619.
		
		\bibitem{torr1998}
		P.~H. Torr, ``Geometric motion segmentation and model selection,''
		\emph{Philosophical Transactions of the Royal Society of London A:
			Mathematical, Physical and Engineering Sciences}, vol. 356, no. 1740, pp.
		1321--1340, 1998.
		
		\bibitem{isack2012}
		H.~Isack and Y.~Boykov, ``Energy-based geometric multi-model fitting,''
		\emph{IJCV}, vol.~97, no.~2, pp. 123--147, 2012.
		
		\bibitem{quigley2009}
		M.~Quigley, K.~Conley, B.~P. Gerkey, J.~Faust, T.~Foote, J.~Leibs, R.~Wheeler,
		and A.~Y. Ng, ``{ROS}: an open-source robot operating system,'' in
		\emph{{ICRA} Workshop on Open Source Software}, May 2009.
		
		\bibitem{furgale2013}
		P.~Furgale, J.~Rehder, and R.~Siegwart, ``Unified temporal and spatial
		calibration for multi-sensor systems,'' in \emph{IROS}, Nov 2013, pp.
		1280--1286.
		
		\bibitem{barfoot2015}
		S.~Anderson and T.~D. Barfoot, ``Full {STEAM} ahead: Exactly sparse {G}aussian
		process regression for batch continuous-time trajectory estimation on
		{SE(3)},'' in \emph{IROS}, Sept 2015, pp. 157--164.
		
		\bibitem{tang2018}
		T.~Y. Tang, D.~J. Yoon, and T.~D. Barfoot, ``A white-noise-on-jerk motion prior
		for continuous-time trajectory estimation on {SE}(3),'' \emph{RA-L}, 2019,
		{T}o appear. arXiv:1809.06518.
		
\end{thebibliography}}
\end{document}